\pdfoutput=1

\documentclass[11pt]{article}

\usepackage[]{ACL2023}

\usepackage{times}
\usepackage{latexsym}

\usepackage[T1]{fontenc}

\usepackage[utf8]{inputenc}

\usepackage{microtype}

\usepackage{inconsolata}
\usepackage{booktabs}
\usepackage{graphicx}
\usepackage{CJKutf8}
\usepackage{cleveref}
\usepackage{multirow}

\title{Extrinsic Evaluation of Machine Translation Metrics}

\author{Nikita Moghe \and Tom Sherborne \and Mark Steedman \and Alexandra Birch \\
    School of Informatics, University of Edinburgh \\ \medskip
   \texttt{\{nikita.moghe, tom.sherborne, a.birch\}@ed.ac.uk} , \texttt{steedman@inf.ed.ac.uk}}

\begin{document}
\setlength{\abovedisplayskip}{2pt}
\setlength{\belowdisplayskip}{2pt}

\maketitle
\begin{abstract}
Automatic machine translation (MT) metrics are widely used to distinguish the 
quality of  machine translation systems across large test sets (i.e., system-level evaluation). 
However, it is unclear if automatic metrics can reliably distinguish good translations from bad at the sentence level (i.e., segment-level evaluation). We investigate how useful MT metrics are at detecting segment-level quality by correlating metrics with the translation utility for downstream tasks.
We evaluate the segment-level performance of widespread MT metrics (chrF, COMET, BERTScore, \textit{etc.}) on three downstream cross-lingual tasks (dialogue state tracking, question answering, and semantic parsing). For each task, we have access to a monolingual task-specific model and a translation model. We calculate the correlation between the metric's ability to predict a good/bad translation with the success/failure on the final task for machine-translated test sentences. 
Our experiments demonstrate that all metrics exhibit negligible correlation with the extrinsic evaluation of downstream outcomes. We also find that the scores provided by neural metrics are not interpretable, in large part due to having undefined ranges.
We synthesise our analysis into recommendations for future MT metrics to produce labels rather than scores for more informative interaction between machine translation and multilingual language understanding.
\end{abstract}
\section{Introduction}
Although machine translation (MT) is typically seen as a standalone application, in recent years MT models have been more frequently deployed as a component of a complex NLP platform delivering multilingual capabilities such as cross-lingual information retrieval \citep{zhang-etal-2022-evaluating} or automated multilingual customer support \citep{gerz-etal-2021-multilingual}. When an erroneous translation is generated by the MT systems, it may add new errors in the task pipeline leading to task failure and poor user experience. For example, consider the user's request in Chinese
\begin{CJK*}{UTF8}{gbsn} 剑桥有牙买加菜吗？\end{CJK*}
(``\textit{Is there any good Jamaican food in Cambridge?}'') machine-translated into English as \textit{``Does Cambridge have a good meal in Jamaica?''}. The model will erroneously consider ``Jamaica'' as a location, instead of cuisine, and prompt the search engine to look up restaurants in Jamaica \footnote{Example from the Multi$^2$WoZ dataset \citep{hung-etal-2022-multi2woz}}. To avoid this \textit{breakdown}, it is crucial to detect an incorrect translation before it causes further errors in the task pipeline. 

One way to approach this \textit{breakdown detection} is using segment-level scores provided by MT metrics. Recent MT metrics have demonstrated 
high correlation with human judgements at the system level for some language pairs \citep{ma-etal-2019-results}. These metrics are potentially capable of identifying subtle differences between MT systems that emerge over a relatively large test corpus. These metrics are also evaluated on respective correlation with human judgements at the segment level, however, there is a considerable performance penalty \citep{ma-etal-2019-results, freitag-etal-2021-results}. Segment-level evaluation of MT is indeed more difficult and even humans have low inter-annotator agreement on this task \citep{popovic-2021-agree}. Despite MT systems being a crucial intermediate step in several applications, characterising the behaviour of these metrics under task-oriented evaluation has not been explored. 
%

In this work, we provide a complementary evaluation of MT metrics. We focus on the segment-level performance of metrics, and we evaluate their performance extrinsically, by correlating each with the outcome of downstream tasks with respective, reliable accuracy metrics. 
We assume access to a parallel task-oriented dataset, a task-specific monolingual model, and a translation model that can translate from the target language into the language of the monolingual model. We consider the \textit{Translate-Test} setting --- where at test time, the examples from the test language are translated to the task language for evaluation. We use the outcomes of this extrinsic task to construct a breakdown detection benchmark for the metrics. 

We use dialogue state tracking, semantic parsing, and extractive question answering as our extrinsic tasks. We evaluate nine metrics consisting of string overlap metrics, embedding-based metrics, and metrics trained using scores from human evaluation of MT. Surprisingly, we find our setup challenging for all existing metrics; demonstrating poor capability in discerning good and bad translations across tasks. We present a comprehensive analysis of the failure of the metrics through quantitative and qualitative evaluation. 


Our contributions are summarised as follows: \\
   1) We derive a new \textbf{breakdown detection task}, for evaluating MT metrics, measuring how indicative segment-level scores are for downstream performance of an extrinsic cross-lingual task (\Cref{sec:methods}). We evaluate nine metrics on three extrinsic tasks covering 39 unique language pairs. The task outputs, the breakdown detection labels, and metric outputs are publicly available.\\ \footnote{\url{https://huggingface.co/datasets/uoe-nlp/extrinsic_mt_eval}}
    2) We show that segment-level scores, from these metrics, have \textbf{minimal correlation with extrinsic task performance} (\Cref{sec:main-results}). Our results indicate that these scores are uninformative at the segment level (\Cref{sec:threshold}) --- clearly demonstrating a serious deficiency in the best contemporary MT metrics. In addition, we find variable task sensitivity to different MT errors (\Cref{sec:qualitative_evaluation}). \\
    3) We propose \textbf{recommendations} on developing MT metrics to produce useful segment-level output by predicting labels instead of scores and suggest reusing existing post-editing datasets and explicit error annotations (See \Cref{sec:recommendatios}).

\section{Related Work}
Evaluation of machine translation has been of great research interest across different communities \citep{nakazawa-etal-2022-overview, fomicheva-etal-2021-eval4nlp}. Notably, the Conference on Machine Translation (WMT) has been organising annual shared tasks on automatic MT evaluation since 2006 \citep{koehn-monz-2006-manual, freitag-etal-2021-results} that invites metric developers to evaluate their methods on outputs of several MT systems. Metric evaluation typically includes a correlation of the scores with human judgements collected for the respective translation outputs. But, designing such guidelines is challenging \citep{mathur-etal-2020-tangled}, leading to the development of several different methodologies and analyses over the years.

The human evaluation protocols include general guidelines for fluency, adequacy and/or comprehensibility \citep{white-etal-1994-arpa} on continuous scales \citep{koehn-monz-2006-manual, graham-etal-2013-continuous} (direct assessments) or fine-grained annotations of MT errors \citep{freitag-etal-2021-experts, freitag-etal-2021-results} based on error ontology like Multidimensional Quality Metrics (MQM) \citep{lommel2014} or rank outputs from different MT systems for the same input \citep{vilar-etal-2007-human}. Furthermore, the best way to compare MT scores with their corresponding judgements is also an open question 
\citep{callison-burch-etal-2006-evaluating, bojar-etal-2014-findings, bojar-etal-2017-results}. The new metrics claim their effectiveness by comparing their performance with competitive metrics on the latest benchmark.


\begin{figure*}
    \centering
     \includegraphics[width=\textwidth]{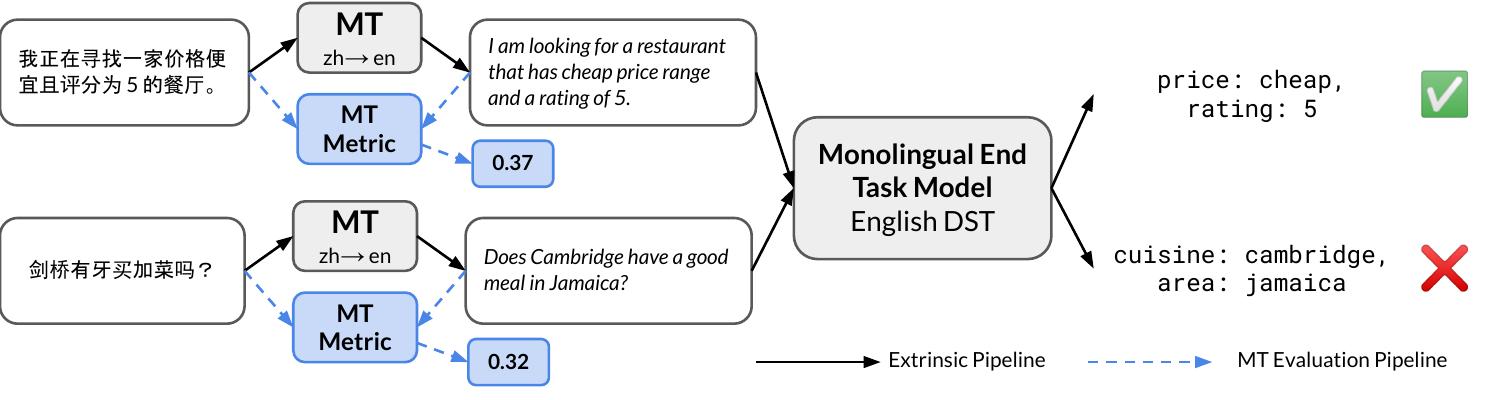}
    \caption{The meta-evaluation pipeline. The predictions for the extrinsic task in the test language (Chinese, ZH) are obtained using the \textit{Translate-Test} setup --- the test language is translated into the task language (English, EN) before passing to the task-specific model. The input sentence (ZH) and the corresponding translations (EN) are evaluated with a metric of interest. The metric is evaluated based on the correlation of its scores with the predictions of the end task.}
    \label{fig:pipeline}
    \vspace{-2ex}
\end{figure*}





The progress and criticism of MT evaluation are generally documented in a metrics shared task overview \citep{callison-burch-etal-2007-meta}. For example, \citet{stanojevic-etal-2015-results} highlighted the effectiveness of neural embedding-based metrics; \citet{ma-etal-2019-results} show that metrics struggle on segment-level performance despite achieving impressive system-level correlation; \citet{mathur-etal-2020-results} investigate how different metrics behave under different domains.  In addition to these overviews, \citet{mathur-etal-2020-tangled} show that meta-evaluation regimes were sensitive to outliers and minimal changes in evaluation metrics are insufficient to claim metric efficacy. \citet{kocmi-etal-2021-ship} conducted a comprehensive evaluation effort to identify which metric is best suited for pairwise ranking of MT systems. \citet{guillou-hardmeier-2018-automatic} look at a specific phenomenon of whether metrics are capable of evaluating translations involving pronominal anaphora. Recent works have also criticised individual metrics such as COMET \citep{amrhein-sennrich-2022-identifying} and BERTScore \citep{hanna-bojar-2021-fine}.

These works draw their conclusions based on some comparison with human judgement or on specific pitfalls of individual metrics. Our work focuses on the usability of the metrics as solely judged on their ability to predict downstream tasks where MT is an intermediate step (with a primary emphasis on segment-level performance). Task-based evaluation has been well studied (\citet{DBLP:books/sp/JonesG96, laoudi-etal-2006-task, zhang-etal-2022-evaluating}, \textit{inter alia}) but limited to evaluating MT systems rather than MT metrics. Closer to our work is \citet{scarton-etal-2019-estimating, zouhar-etal-2021-neural} which proposes MT evaluation as ranking translations based on the time to post-edit model outputs. 
We borrow the term of \textit{breakdown detection} from \citet{Martinovsky2006TheEI} that proposes breakdown detection for dialogue systems to detect unnatural responses.

\section{Methodology}
\label{sec:methods}

Our aim is to determine how reliable MT metrics are for predicting success on downstream tasks. 
Our setup uses a monolingual model (e.g., a dialogue state tracker) trained on a \textit{task language} and parallel test data from multiple languages. We use MT to translate a test sentence (from a \textit{test language} to the \textit{task language}) and then infer a label for this example using the monolingual model. If the model predicts a correct label for the parallel \textit{task language} input but an incorrect label for the translated \textit{test language} input, then we have observed a \textit{breakdown} due to a material error in the translation pipeline. We then study if the metric could predict if the translation is suitable for the end task.  We refer to \Cref{fig:pipeline} for an illustration. We frequently use the terms \textit{test language} and \textit{task language} to avoid confusion with the usage of \textit{source language} and \textit{target language} in the traditional machine translation setup. In \Cref{fig:pipeline}, the task language is English and the test language is Chinese. We now describe our evaluation setup and the metrics under investigation.

\subsection{Setup}

For all the tasks described below, we first train a model for the respective tasks on the monolingual setup. We evaluate the task language examples on each task and capture the monolingual predictions of the model. We consider the \textit{Translate-Test} paradigm \citep{hu-etal-xtreme}, we translate the examples from each test language into the task language. The generated translations are then fed to the task-specific monolingual model. We use either (i) OPUS translation models \citep{tiedemann-thottingal-2020-opus}, (ii) M2M100 translation \citep{JMLR:v22:20-1307} or (iii) translations provided by the authors of respective datasets. Note that the examples across all the languages are parallel and we therefore always have access to the correct label for a translated sentence. We obtain the predictions for the translated data to construct a breakdown detection benchmark for the metrics.

We consider only the subset of examples in the test language which were correctly predicted in the task language to avoid errors that arise from extrinsic task complexity. Therefore, all incorrect extrinsic predictions for the test language in our setup arise from erroneous translation. This isolates the extrinsic task failure as the fault of \textit{only} the MT system. We use these predictions to build a binary classification benchmark---all target language examples that are correctly predicted in the extrinsic task receive a positive label (no breakdown) while the incorrect predictions receive a negative label (breakdown).

We consider the example from the test language as \textit{source}, the corresponding machine translation as \textit{hypothesis} and the human reference from the task language as \textit{reference}. Thus, in \Cref{fig:pipeline}, the source is  \begin{CJK*}{UTF8}{gbsn} 剑桥有牙买加菜吗？\end{CJK*}, the hypothesis is ``Does Cambridge have a good meal in Jamaica", and the reference will be ``Is there any good Jamaican food in Cambridge". These triples are then scored by the respective metrics. After obtaining the segment-level scores for these triples, we define a threshold for the scores, thus turning metrics into classifiers. For example, if the threshold for the metric in \Cref{fig:pipeline} is 0.5, it would mark both examples as bad translations. We plot a histogram over the scores with ten bins for every setup and select the interval with the highest performance on the development set as a threshold. The metrics are then evaluated on how well their predictions for a good/bad translation correlate with the breakdown detection labels.

\subsection{Tasks}
We choose tasks that contain outcomes belonging to a small set of labels, unlike natural language generation tasks which have a large solution space. This discrete nature of the outcomes allows us to quantify the performance of MT metrics based on standard classification metrics. The tasks also include varying types of textual units:~utterances, sentences, questions, and paragraphs, allowing a comprehensive evaluation of the metrics. 

\subsubsection{Semantic Parsing (SP)}

Semantic parsing transforms natural language utterances into logical forms to express utterance semantics in some machine-readable language. The original ATIS study \citep{hemphill-etal-1990-atis} collected questions about flights in the USA with the corresponding SQL to answer respective questions from a relational database. We use the MultiATIS++SQL dataset from \citet{sherborne-lapata-2022-zero} comprising gold parallel utterances in English, French, Portuguese, Spanish, German and Chinese (from \citet{xu-etal-2020-end}) paired to executable SQL output logical forms (from \citet{iyer-etal-2017-learning}). The model follows \citet{10.1162/tacl_a_00533}, as an encoder-decoder Transformer model based on mBART50 \citep{tang-etal-2021-multilingual}. The parser generates valid SQL queries and performance is measured as exact-match \textit{denotation accuracy}---the proportion of output queries returning identical database results relative to gold SQL queries.

\subsubsection{Extractive Question Answering (QA)}
The task of extractive question answering is predicting a span of words from a paragraph corresponding to the question. We use the XQuAD dataset \citep{artetxe-etal-2020-cross} for evaluating extractive question answering.  The XQuAD dataset was obtained by professionally translating examples from the development set of English SQuAD dataset \citep{rajpurkar-etal-2016-squad} into ten languages:~Spanish, German, Greek, Russian, Turkish, Arabic, Vietnamese, Thai, Chinese, and Hindi. We use the publicly available question answering model that fine-tunes RoBERTa \citep{DBLP:journals/corr/abs-1907-11692} on the SQuAD training set. 
We use the \textit{Exact-Match} metric, i.e., the model's predicted answer span exactly matches the gold standard answer span; for the breakdown detection task. The metrics scores are produced for the question and the context. A translation is considered to be faulty if either of the scores falls below the chosen threshold for every metric.

\subsubsection{Dialogue State Tracking (DST)}
In the dialogue state tracking task, a model needs to map the user's goals and intents in a given conversation to a set of slots and values,~known as a \textit{dialogue state}, based on a pre-defined ontology. MultiWoZ 2.1 \citep{eric-etal-2020-multiwoz} is a popular dataset for examining the progress in dialogue state tracking which consists of multi-turn conversations in English spanning across 7 domains. We consider the Multi$^{2}$WoZ dataset \citep{hung-etal-2022-multi2woz} where the development and test set have been professionally translated into German, Russian, Chinese, and Arabic from the MultiWoZ 2.1 dataset.  We use the dialogue state tracking model trained on the English dataset by \citet{lee-etal-2019-sumbt}. We consider the \textit{Joint Goal Accuracy} where the inferred label is correct only if the predicted dialogue state is exactly equal to the ground truth to provide labels for the breakdown task. We use oracle dialogue history and the metric scores are produced only for the current utterance spoken by the user.

\subsection{Metrics}
We describe the metrics based on their design principles:~derived from the surface level token overlap, embedding similarity, and neural metrics trained using WMT data. We selected the following metrics as they are the most studied, frequently used, and display a varied mix of design principles. 

\subsubsection{Surface Level Overlap}
\textbf{BLEU} \citep{papineni-etal-2002-bleu} is a string-matching metric that compares the token-level n-grams of the hypothesis with the reference translation.  BLEU is computed as a precision score weighted by a brevity penalty. We use sentence-level BLEU in our experiments. 

\noindent \textbf{chrF} \citep{popovic-2017-chrf} computes a character n-gram F-score based on the overlap between the hypothesis and the reference.

\subsubsection{Embedding Based}
\noindent \textbf{BERTScore} \citep{DBLP:conf/iclr/ZhangKWWA20} uses contextual embeddings from pre-trained language models to compute the similarity between the tokens in the reference and the generated translation using cosine similarity. The similarity matrix is used to compute precision, recall, and F1 scores.

\subsubsection{Trained on WMT Data}
WMT organises an annual shared task on developing MT models for several categories in machine translation \citep{akhbardeh-etal-2021-findings}. Human evaluation of the translated outputs from the participating machine translation models is often used to determine the best-performing MT system. In recent years, this human evaluation has followed two protocols:~(i) Direct Assessment (DA) \citep{graham-etal-2013-continuous}: where the given translation is rated from 0 to 100 based on the perceived translation quality and (ii) Expert based evaluation where the translations are evaluated by professional translators with explicit error listing based on the Multidimensional Quality Metrics (MQM) ontology. MQM ontology consists of a hierarchy of errors and translations are penalised based on the severity of errors in this hierarchy. These human evaluations are then used as training data for building new MT metrics. 


\noindent \textbf{COMET metrics}:
Cross-lingual Optimized Metric for Evaluation of Translation (COMET) \citep{rei-etal-2020-comet} 
uses a cross-lingual encoder (XLM-R \citep{conneau-etal-2020-unsupervised}) and pooling operations to predict score of the given translation. Representations for the source, hypothesis, and reference (obtained using the encoder) are combined and passed through a feedforward layer to predict a score. These metrics use a combination of WMT evaluation data across the years to produce different metrics. In all the variants, the MQM scores and DA scores are normalised to z-scores to reduce the effect of outlier annotations. 
 \\
\noindent \textbf{COMET-DA} uses direct assessments from 2017 to 2019 as training data while \textbf{COMET-MQM} uses direct assessments from 2017 to 2021 as training data. This metric is then fine-tuned with MQM data from \citet{freitag-etal-2021-experts}.

\noindent \textbf{UniTE metrics} \citep{wan-etal-2022-unite}, Unified Translation Evaluation, is another neural translation metric that proposes a multi-task setup for the three strategies of evaluation:~source-hypothesis, source-hypothesis-reference, and reference-hypothesis in a single model. The pre-training stage involves training the model with synthetic data constructed using a subset of WMT evaluation data. Fine-tuning uses novel attention mechanisms and aggregate loss functions to facilitate the multi-task setup. 

All the above reference-based metrics have their corresponding reference-free versions which use the same training regimes but exclude encoding the reference. We refer to them as COMET-QE-DA, COMET-QE-MQM, and UniTE-QE respectively. COMET-QE-DA in this work uses DA scores from 2017 to 2020. We list the code sources of these metrics in \Cref{app:metric}.

\subsection{Metric Evaluation}
The meta-evaluation for the above metrics uses the breakdown detection benchmark. As the class distribution changes depending on the task and the language pair, we require an evaluation that is robust to class imbalance. We consider using macro-F1 and Matthew's Correlation Coefficient (MCC) \citep{MATTHEWS1975442} on the classification labels. The range of macro-F1 is from 0 to 1 with equal weight to positive and negative classes. 
We include MCC to interpret the MT metric's standalone performance for the given extrinsic task.  The range of MCC is between -1 to 1. An MCC value near 0 indicates no correlation with the class distribution. Any MCC value between 0 and 0.3 indicates negligible correlation, 0.3 to 0.5 indicates low correlation.

\section{Results}
\label{sec:results}
\begin{table*}
\small
\centering
\begin{tabular}{@{}lcccccc@{}}
\toprule
Metric & \multicolumn{2}{l}{Semantic Parsing} & \multicolumn{2}{l}{Question Answering} & \multicolumn{2}{l}{Dialogue State Tracking} \\ 
 & F1 & MCC & F1 & MCC & F1 & MCC \\ \midrule
Random  & 0.453 & -0.034 & 0.496 & 0.008  & 0.493 & 0.008\\ \midrule
BLEU  & 0.580 & 0.179 & 0.548 & 0.121 & 0.529 & 0.082 \\
chrF  & 0.609 & 0.234 & 0.554 & 0.127 & 0.508 & 0.067\\ \midrule
BERTScore  & 0.590 & 0.205 & 0.555 & 0.127 & 0.505 & 0.071 \\
COMET-DA  & 0.606 & 0.228 & 0.562 & 0.137 & 0.608 & 0.244\\
COMET-MQM  & 0.556 & 0.132 & 0.387 & 0.027 & 0.597 & 0.204\\
UniTE  & 0.600 & 0.225 & 0.375 & 0.012 & 0.620 & 0.262 \\ \midrule
COMET-QE-DA  & 0.556 & 0.135 & 0.532 & 0.100 & 0.561 & 0.145\\
COMET-QE-MQM  & 0.597 & 0.211 & 0.457 & 0.033 & 0.523 & 0.094 \\
UniTE-QE  & 0.567 & 0.155 & 0.388 & 0.032 & 0.587 & 0.192 \\ \midrule
Ensemble  & 0.620 & 0.251 & 0.577 & 0.168 & 0.618 & 0.248 \\ \bottomrule
\end{tabular}
\caption{Performance of MT metrics on the classification task  for extrinsic tasks  Parsing (MultiATIS++SQL), Question Answering  (XQuad) using an English-trained question answering system, and Dialogue State Tracking (Multi$^2$WoZ) using an English-trained state tracker. Reported Macro F1 scores and MCC scores quantify if the metric detects a breakdown for the extrinsic task. Metrics have a negligible correlation with the outcomes of the end task. MCC and F1 are average over respective language pairs}
\label{tab:main}
\vspace{-2ex}
\end{table*}

We report the aggregated results for semantic parsing, question answering, and dialogue state tracking in \Cref{tab:main} with fine-grained results in \Cref{app:task}. We use a random baseline for comparison which assigns the positive and negative labels with equal probability. 
\subsection{Performance on Extrinsic Tasks}
\label{sec:main-results}
We find that almost all metrics perform above the random baseline on the macro-F1 metric. We use MCC to identify if this increase in macro-F1 makes the metric usable in the end task. Evaluating MCC, we find that all the metrics show negligible correlation across all three tasks. Contrary to trends where neural metrics are better than metrics based on surface overlap \citep{freitag-etal-2021-results}, we find this breakdown detection to be difficult irrespective of the design of the metric. We also evaluate an ensemble with majority voting of the predictions from the top three metrics per task. Ensembling provides minimal gains suggesting that metrics are making similar mistakes despite varying properties of the metrics.



Comparing the reference-based versions of trained metrics (COMET-DA, COMET-MQM, UniTE) with their reference-free quality estimation (QE) equivalents, we observe that reference-based versions perform better, or are competitive to, their reference-free versions for the three tasks. We also note that references are unavailable when the systems are in production, hence reference-based metrics are unsuitable for realistic settings. We discuss alternative ways of obtaining references in \Cref{sec:rtt-augment}. 

Between the use of MQM-scores and DA-scores during fine-tuning COMET variants, we find that both COMET-QE-DA and COMET-DA are strictly better than COMET-QE-MQM and COMET-MQM for question answering and dialogue state tracking respectively, with no clear winner for semantic parsing (See \Cref{app:task}).

The results on per-language pair in \Cref{app:task} suggest that no specific language pairs stand out as easier/harder across tasks. As this performance is already poor, we cannot verify if neural metrics can generalise in evaluating language pairs unseen during training.

\begin{table*}
\small
\centering
\begin{tabular}{@{}lrll@{}}
\toprule
Task &
  \multicolumn{1}{l}{\begin{tabular}[c]{@{}l@{}}Errors by the\\ Extrinsic model\end{tabular}} &
  False Positive &
  False Negative \\ \midrule
SP &
  25\% &
  mistranslation (90\%), omission(10\%) &
  \begin{tabular}[c]{@{}l@{}}mistranslation (25.7\%), fluency (20\%),\\  omission (5.7\%), no error (48.6\%)\end{tabular} \\
QA &
  20\% &
  \begin{tabular}[c]{@{}l@{}}mistranslation (60\%), omission(8.6\%), \\ addition (5.7\%), fluency (20\%),\\ undertranslation (2.9\%), untranslated (2.9\%)\end{tabular} &
  \begin{tabular}[c]{@{}l@{}}mistranslation (18\%), fluency (22\%), \\ addition (2\%), no error (54\%)\end{tabular} \\
DST &
  5\% &
  mistranslation (100\%) &
  \begin{tabular}[c]{@{}l@{}}omission (26\%), mistranslation (1\%),\\  no error (73\%)\end{tabular} \\ \bottomrule
\end{tabular}
\caption{The proportion of the different types of errors erroneously detected and undetected by COMET-DA for languages mentioned in Section ~\ref{sec:qualitative_evaluation}. False positives and false negatives are computed by excluding the examples where the extrinsic task model was at fault. }
\label{tab:error-quant}
\end{table*}

\begin{figure}
    \centering
    \includegraphics[scale=0.3]{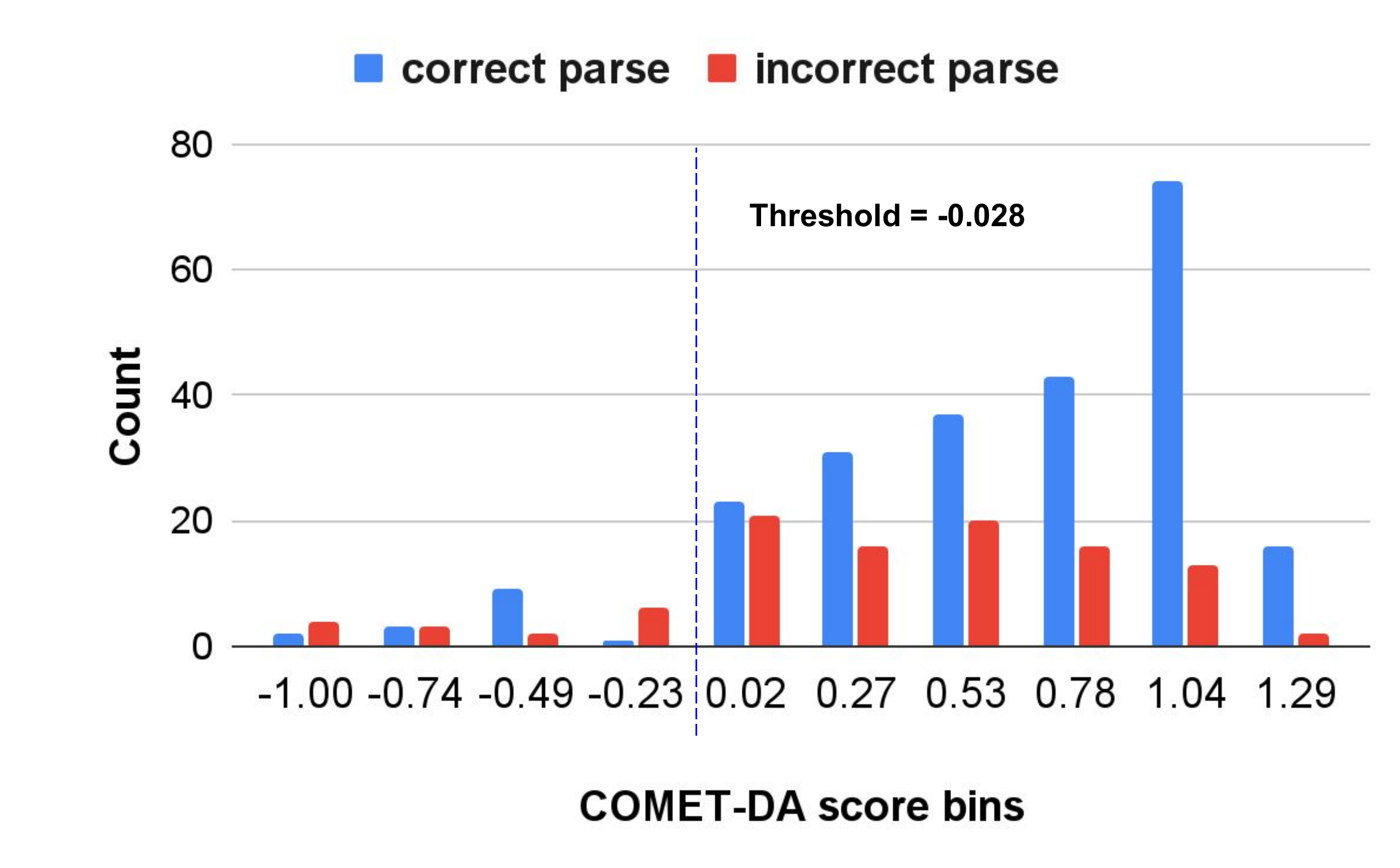}
    \caption{Graph of predictions by COMET-DA (threshold: -0.028), categorised by the metric scores in ten intervals. Task: Semantic Parsing with English parser and test language is Chinese. The bars indicate the count of examples with incorrect parses (red) and correct parses (blue) assigned the scores for the given ranges. }
    \label{fig:qual-ex}
\end{figure}

\noindent \textbf{Case Study}: We look at Semantic Parsing with an English-trained parser tested with Chinese inputs for our case study with the well-studied COMET-DA metric. We report the number of correct and incorrect predictions made by COMET-DA across ten equal ranges of scores in \Cref{fig:qual-ex}. The bars labelled on the x-axis indicate the end-point of the interval i.e., the bar labelled -0.74 contains examples that were given scores between -1.00 and -0.74.

First, we highlight that the threshold is -0.028, counter-intuitively suggesting that even some correct translations receive a negative score. We expected the metric to fail in the regions around the threshold as those represent strongest confusion. For example,
\begin{CJK*}{UTF8}{gbsn}
``周日 下午 从 迈阿密 飞往 克利夫兰''
\end{CJK*} is correctly translated as ``Sunday afternoon from Miami to Cleveland'' yet the metric assigns it a score of -0.1. However, the metric makes mistakes throughout the bins. For example, \begin{CJK*}{UTF8}{gbsn}
``我需要预订一趟 联合航空 下 周六 的从 辛辛那提 飞往 纽约市 的航班''
\end{CJK*} 
is translated as ``I need to book a flight from Cincinnati to New York City next Saturday.'' and loses the crucial information of ``United Airlines''; yet it is assigned a high score of 0.51. This demonstrates that the metric possesses a limited perception of a good or bad translation for the end task. 

We suspect this behaviour is due to the current framework of MT evaluation. The development of machine translation metrics largely caters towards the intrinsic task of evaluating the quality of a translated text in the target language. The severity of a translation error is dependent on the guidelines released by the organisers of the WMT metrics task or the design choices of the metric developers. Our findings agree with \citet{zhang-etal-2022-evaluating}  that different downstream tasks will demonstrate varying levels of sensitivity to the same machine translation errors.

\subsection{Qualitative Evaluation}

\label{sec:qualitative_evaluation}
To quantify detecting which translation errors are most crucial to the respective extrinsic tasks, we conduct a qualitative evaluation of the MT outputs and task predictions. We annotate 50 false positives and 50 false negatives for test languages Chinese (SP), Hindi (QA), and Russian (DST) respectively.  The task language is English. We annotate the MT errors (if present) in these examples based on the MQM ontology. We tabulate these results in \Cref{tab:error-quant} using COMET-DA for these analyses.  

Within the false negatives, a majority of the errors (>48\%) are due to the metric's inability to detect translations containing synonyms or paraphrases of the references as valid translations. Further, omission errors detected by the metric are not crucial for DST as these translations often exclude pleasantries. Similarly, errors in fluency are not important for both DST and SP but they are crucial for QA as grammatical errors in questions produce incorrect answers. Mistranslation of named entities (NEs), especially which lie in the answer span, is a false negative for QA since QA models find the answer by focusing on the words in the context surrounding the NE rather than the error in that NE. Detecting mistranslation in NEs is crucial for both DST and SP as this error category dominates the false positives. A minor typo of \textit{Lester} instead of \textit{Leicester} marks the wrong location in the dialogue state which is often undetected by the metric. Addition and omission errors are also undetected for SP while mistranslation of reservation times is undetected for DST.




We also find that some of the erroneous predictions can be attributed to the failure of the extrinsic task model than the metric. For example, the MT model uses an alternative term of  \textit{direct} instead of \textit{nonstop} while generating the translation for the reference ``show me nonstop flights from montreal to orlando''. The semantic parser fails to generalise despite being trained with mBART50 to ideally inherit some skill at disambiguiting semantically similar phrases. This error type accounts for 25\% for SP, 20\% for QA and 5\% in DST of the total annotated errors. We give examples in \Cref{app:error}.


\subsection{Finding the Threshold}

\label{sec:threshold}
    

\begin{table}[t]
\small
\centering
\resizebox{\columnwidth}{!}{%
\begin{tabular}{@{}cccc@{}}
\toprule
Extrinsic Task & SP & QA & DST \\ \midrule
BLEU & 15.5   $\pm$ 08.8 & 16.1   $\pm$ 04.9 & 20.0  $\pm$ 0.00 \\ 
chrF & 44.0   $\pm$ 13.7 & 53.9   $\pm$ 07.8 & 30.7  $\pm$ 0.45 \\ \midrule
BERTScore & 0.50  $\pm$ 0.21 & 0.54  $\pm$ 0.08 & 0.39 $\pm$ 0.21 \\
COMET-DA & 0.21  $\pm$ 0.35 & 0.30  $\pm$ 0.23 & 0.58 $\pm$ 0.08 \\
COMET-MQM & 0.03  $\pm$ 0.01 & 0.06  $\pm$ 0.01 & 0.02 $\pm$ 0.00 \\
UniTE & 0.04  $\pm$ 0.22 & -0.40 $\pm$ 0.38 & -0.01 $\pm$ 0.29 \\ \midrule
COMET-QE-DA & 0.02  $\pm$ 0.07 & 0.02  $\pm$ 0.01 & 0.06 $\pm$ 0.01 \\
COMET-QE-MQM & 0.11  $\pm$ 0.01 & 0.00  $\pm$ 0.04 & 0.03 $\pm$ 0.00 \\
UniTE-QE & -0.01 $\pm$ 0.22 & -0.24 $\pm$ 0.13 & 0.11 $\pm$ 0.18 \\ \bottomrule
\end{tabular}%
}
\caption{Mean and Standard Deviation of the best threshold on the development set for all the language pairs in the respective extrinsic tasks. The thresholds are inconsistent across language pairs and tasks for both bounded and unbounded metrics. }
\label{tab:threshold}
\end{table}

Interpreting system-level scores provided by automatic metrics requires additional context such as the language pair of the machine translation model or another MT system for comparison \footnote{\url{https://github.com/Unbabel/COMET/issues/18}}. In this classification setup, we rely on interpreting the segment-level score to determine whether the translation is suitable for the downstream task. We find that choosing the right threshold to identify translations requiring correction is not straightforward. Our current method to obtain a threshold relies on validating candidate thresholds on the development set and selecting an option with the best F1 score. These different thresholds are obtained by plotting a histogram of scores with ten bins per task and language pair. 

We report the mean and standard deviation of best thresholds for every language pair for every metric in \Cref{tab:threshold}. Surprisingly, the thresholds are inconsistent and biased for bounded metrics:~BLEU (0--100), chrF (0--100), and BERTScore (0--1). The standard deviations across the table indicate that the threshold varies greatly across language pairs.  We find that thresholds of these metrics are also not transferable across tasks. COMET metrics, except COMET-DA, have lower standard deviations. By design, the range of COMET metrics in this work is unbounded. However, as discussed in the theoretical range of COMET metrics \footnote{\url{https://unbabel.github.io/COMET/html/faqs.html}}, empirically, the range for COMET-MQM lies between -0.2 to 0.2, questioning whether lower standard deviation is an indicator of threshold consistency. Some language pairs within the COMET metrics have negative thresholds. We also find that some of the use cases under the UniTE metrics have a mean negative threshold, indicating that good translations can have negative UniTE scores. Similar to \citet{https://doi.org/10.48550/arxiv.2209.14172}, we suggest that the notion of negative scores for good translations, only for certain language pairs, is counter-intuitive as most NLP metrics tend to produce positive scores. 

Thus, we find that both bounded and unbounded metrics discussed here do not provide segment-level scores whose range can be interpreted meaningfully across tasks and language pairs.


%

\subsection{Reference-based Metrics in an Online Setting}

\label{sec:rtt-augment}
In an online setting, we do not have access to references at test time. To test the effectiveness of reference-based methods here, we consider translating the translation back into the test language. For example, for an \textit{en} parser, the test language \textit{$ti_{zh}$} is translated into \textit{$mt_{en}$} and then translated back to Chinese as \textit{$mt_{zh}$}. The metrics now consider \textit{$mt_{en}$} as source, \textit{$mt_{zh}$} as hypothesis, and \textit{$ti_{zh}$} as the reference. We generate these new translations using the mBART50 translation model \citep{tang-etal-2021-multilingual} and report the results in \Cref{tab:rtt-all}.
\begin{table}[t]
\centering
\small
\begin{tabular}{@{}clll@{}}
\toprule
Metric    & SP    & QA    & DST   \\ \midrule
BLEU      & 0.003 & 0.013 & 0.050 \\
chrF      & 0.018 & 0.021 & 0.055 \\ \midrule
BERTScore & 0.028 & 0.065 & 0.036 \\
COMET-DA  & 0.071 & 0.085 & 0.083 \\
COMET-MQM & 0.080 & 0.019 & 0.116 \\ 
UniTE &  0.225 & 0.056 & 0.193 \\
\bottomrule
\end{tabular}
\caption{MCC scores of reference based metrics with pseudo references when gold references are unavailable at  test time. Performance is worse than metrics with oracle references and reference-free metrics (Table \ref{tab:main})}
\label{tab:rtt-all}
\vspace{-2ex}
\end{table}

Compared to the results in \Cref{tab:main}, there is a further drop in performance across all the tasks and metrics. The metrics also perform worse than their reference-free counterparts. The second translation is likely to add additional errors to the existing translation. This cascading of errors confuses the metric and it can mark a perfectly useful translation as a breakdown.  The only exception is that of the UniTE metric which has comparable performance (but overall poor) due to its multi-task setup.

\section{Recommendations}
\label{sec:recommendatios}
Our experiments suggest that evaluating MT metrics on the segment level for extrinsic tasks has considerable room for improvement. We propose recommendations based on our observations:

\textbf{Prefer MQM for Human Evaluation of MT outputs}: 
We  reinforce the proposal of using the MQM scoring scheme with expert annotators for evaluating MT outputs in line with \citet{freitag-etal-2021-experts}.  As seen in \Cref{sec:qualitative_evaluation}, different tasks have varying tolerance to different MT errors. With explicit errors marked per MT output, future classifiers can be trained on a subset of human evaluation data containing errors most relevant to the downstream application.

\textbf{MT Metrics Could Produce Labels over Scores}: The observations from \Cref{sec:qualitative_evaluation} and \Cref{sec:threshold} suggest that interpreting the quality of the produced MT translation based on a number is unreliable and difficult. We recommend exploring whether segment-level MT evaluation can be approached as an error classification task instead of regression. Specifically, whether the words in the source/hypothesis can be tagged with explicit error labels.  Resorting to MQM-like human evaluation will result in a rich repository of human evaluation based on an ontology of errors and erroneous spans marked across the source and hypothesis \citep{freitag-etal-2021-experts}. Similarly, the post-editing datasets  (\citet{scarton-etal-2019-estimating, fomicheva-etal-2022-mlqe} , \textit{inter alia}) also provide a starting point. Recent exploration in this direction are the works by \citet{MATESE:WMT22, COMET:WMT22} that treat MT evaluation as a sequence-tagging problem by labelling the errors in an example. Such metrics can also be used for intrinsic evaluation by assigning weights to the labels to produce a weighted score. 


\textbf{Add Diverse References During Training}: From \Cref{sec:qualitative_evaluation}, we find that both the neural metric and the task-specific model are not robust to paraphrases.  We also recommend the inclusion of diverse references through automatic paraphrasing \citep{bawden-etal-2020-study} or data augmentation during the training of neural metrics. 


\section{Conclusion}
We propose a method for evaluating MT metrics which is reliable at the segment-level and does not depend on human judgements by using correlation MT metrics with the success of extrinsic downstream tasks. We evaluated nine different metrics on the ability to detect errors in generated translations when machine translation is used as an intermediate step for three extrinsic tasks:~Semantic Parsing, Question Answering, and Dialogue State Tracking. We find that segment-level scores provided by all the metrics show negligible correlation with the success/failure outcomes of the end task across different language pairs. We attribute this result to segment scores produced by these metrics being uninformative and that different extrinsic tasks demonstrate different levels of sensitivity to different MT errors. We propose recommendations to predict error types instead of error scores to facilitate the use of MT metrics in downstream tasks. 

\section{Limitations}
As seen in \Cref{sec:qualitative_evaluation}, sometimes the metrics are unnecessarily penalised due to errors made by the end task models. Filtering these cases would require checking every example in every task manually. We hope our results can provide conclusive trends to the metric developers focusing on segment-level MT evaluation.

We included three tasks to cover different types of errors in machine translations and different types of contexts in which an online MT metric is required. Naturally, this regime can be extended to other datasets, other tasks, and other languages \citep{ruder-etal-2021-xtreme, indic-xtreme}. Further, our tasks used stricter evaluation metrics such as exact match. Incorporating information from partially correct outputs is not trivial and will be hopefully addressed in the future.   
We have covered 37 language pairs across the tasks which majorly use English as one of the languages. Most of the language pairs in this study are high-resource languages. Similarly, the examples in multilingual datasets are likely to exhibit \textit{translationese} - unnatural artefacts from the task language present in the test language during manual translation; which tend to overestimate the performance of the various tasks \citep{majewska2022cross, freitag-etal-2020-bleu}. We hope to explore the effect of translationese on MT evaluation \citep{graham-etal-2020-statistical} and extrinsic tasks in future. 
The choice of metrics in this work is not exhaustive and is dependent on the availability and ease of use of the metric provided by the authors. 

\section{Ethics Statement}
This work uses datasets, models, and metrics that are publicly available. Although the scope of this work does not allow us to have an in-depth discussion of biases associated with metrics \citep{amrhein-moghe-guillou:2022:WMT}, we caution the readers of drawbacks of metrics that cause unfair evaluation to marginalised subpopulations which are discovered or yet to be discovered. We will release the translations, metrics scores, and corresponding task outputs for reproducibility.

\section{Acknowledgements}
We thank Barry Haddow for providing us with valuable feedback on setting up this work. We thank Arushi Goel and the attendees at the MT Marathon 2022 for discussions about this work. We thank Ankita Vinay Moghe, Nikolay Bogoychev, and Chantal Amrhein for their comments on the earlier drafts.  We thank the anonymous reviewers for their helpful suggestions. This work was supported in part by the UKRI Centre for Doctoral Training in Natural Language Processing, funded by the UKRI (grant EP/S022481/1) and the University of Edinburgh (Moghe). We also thank Huawei for their support (Moghe). Sherborne gratefully acknowledges the support of the UK Engineering and Physical Sciences Research Council (grant EP/W002876/1).

\bibliography{anthology,custom}
\bibliographystyle{acl_natbib}

\clearpage

\appendix
\section{Language Codes}
Please find the language codes in \Cref{tab:lang-code}.

\begin{table}[]
\small
\centering
\begin{tabular}{@{}llll@{}}
\toprule
Code & Language   & Code & Language   \\ \midrule
en   & English    & el   & Greek      \\
de   & German     & es   & Spanish    \\
zh   & Mandarin Chinese    & hi   & Hindi      \\
fr   & French     & th   & Thai       \\
ar   & Arabic     & tr   & Turkish    \\
ru   & Russian    & vi   & Vietnamese \\
pt   & Portuguese &      &            \\ \bottomrule
\end{tabular}
\caption{Language codes of languages used in this work}
\label{tab:lang-code}
\end{table}
\section{Implementation Details}
\label{app:metric}
We provide the implementation details of metrics and models in \Cref{tab:git-repo}. All models are publicly available and required no training from our side. The metrics BERTScore, COMET family and UniTE family can run on both GPU and CPU. If run on GPU, the metrics run under 5 minutes for a given task and given language pair. No hyperparameters are required. We follow the standard train-dev-test split as released by the authors for DST \citep{hung-etal-2022-multi2woz} and SP \citep{sherborne-lapata-2022-zero}. As no development set is available for the XQuAD dataset, we use the first 200 examples as development set to choose the threshold but report the performance on the full test set.
\begin{table*}
\centering
\tiny
\begin{tabular}{@{}lll@{}}
\toprule
Method & Code & Notes \\ \midrule
\multicolumn{3}{c}{Metrics} \\ \hline
chrF & \url{https://github.com/mjpost/sacrebleu} & Signature: "nrefs:1|case:mixed|eff:no|tok:13a|smooth:exp|version:2.1.0" \\
BLEU & \url{https://github.com/mjpost/sacrebleu} & Signature: "nrefs:1|case:mixed|eff:yes|nc:6|nw:0|space:no|version:2.1.0" \\
BERTScore & \url{https://github.com/Tiiiger/bert\_score} & Model: xlm-roberta-large \\
COMET-DA & \multirow{4}{*}{\url{https://github.com/Unbabel/COMET}} & Model:  wmt20-comet-da \\
COMET-MQM &  & Model:  wmt21-comet-mqm \\
COMET-QE-DA &  & Model:  wmt21-comet-qe-da \\
COMET-QE-MQM &  & Model: wmt21-comet-qe-mqm \\
UniTE & \multirow{2}{*}{\url{https://github.com/NLP2CT/UniTE}} & Model: UniTE-MUP, hparams.src\_ref.yaml \\
UniTE-QE &  & Model: UniTE-MUP, hparams.src.yaml \\ \hline
\multicolumn{3}{c}{Extrinsic Task Models} \\ \hline
SP & \url{https://github.com/tomsherborne/zx-parse} & \\
DST & \url{https://github.com/thu-coai/ConvLab-2} & \\
QA & \url{https://huggingface.co/csarron/roberta-base-squad-v1} & \\ \hline

\end{tabular}
\caption{Metric repositories and versions}
\label{tab:git-repo}
\end{table*}

\section{Errors of COMET-DA}
\label{app:error}
\begin{figure*}
    \centering
    \includegraphics[width=\linewidth]{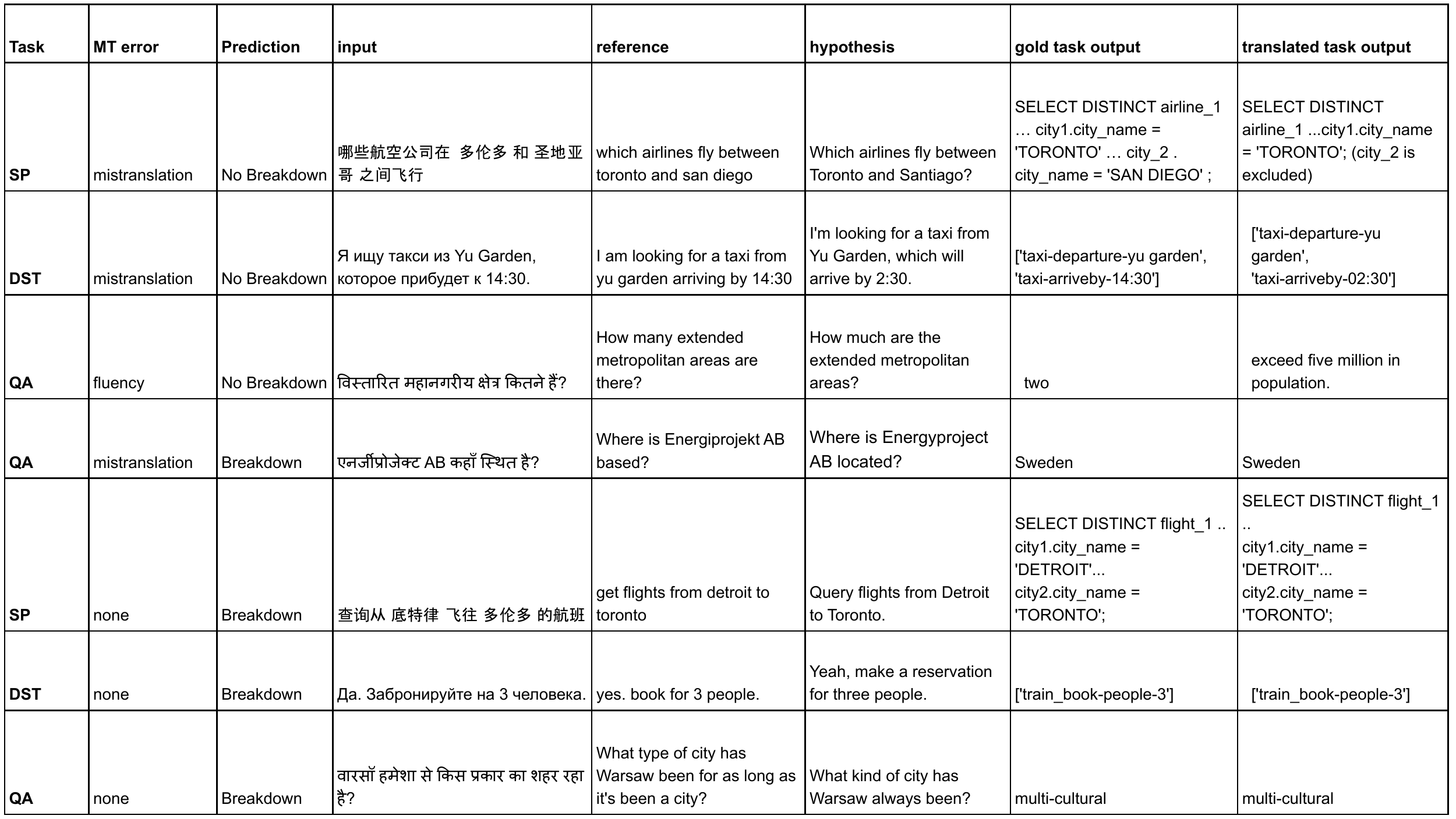}
    \caption{Examples of errors made by COMET-DA}
    \label{fig:errors-qual}
\end{figure*}

The proportion of errors from \Cref{sec:qualitative_evaluation} are listed in \Cref{tab:error-quant}. We also provide error examples in \Cref{fig:errors-qual}.

\section{Task-specific results}
\label{app:task}
We now list the results across every language pair for all the tasks in Tables \cref{tab:dst-main,tab:xquad-main,tab:xquad-F1,tab:parsing-f1,tab:parsing-mcc}. 


\begin{table*}[t]
\small
\centering
\begin{tabular}{@{}ccccccccc@{}}
\toprule
Language & \multicolumn{2}{c}{zh} & \multicolumn{2}{c}{de} & \multicolumn{2}{c}{ar} & \multicolumn{2}{c}{ru} \\
Good / Bad & \multicolumn{2}{c}{1465 / 1796} & \multicolumn{2}{c}{2162 / 1099} & \multicolumn{2}{c}{1744 / 1517} & \multicolumn{2}{c}{1517 / 1744} \\ \midrule
Method & F1 & MCC & F1 & MCC & F1 & MCC & F1 & MCC \\ \midrule
Random & 0.449 & -0.013 & 0.417 & 0.018 & 0.429 & -0.018 & 0.454 & 0.004  \\ \midrule
BLEU & 0.511 & 0.079  & 0.541 & 0.091 & 0.540  & 0.083  & 0.527 & 0.076 \\
chrF &  0.518 & 0.078  & 0.496 & 0.033 & 0.499 & 0.071  & 0.52  & 0.086 \\ \midrule
BERTScore & 0.438 & 0.000      & 0.519 & 0.068 & 0.546 & 0.136  & 0.518 & 0.080 \\
COMET-DA & 0.611 & 0.248  & 0.581 & 0.181 & 0.664 & 0.328  & 0.579 & 0.220 \\
COMET-MQM & 0.594 & 0.201  & 0.574 & 0.165 & 0.625 & 0.255  & 0.598 & 0.196\\
UniTE & 0.642 & 0.285  & 0.572 & 0.164 & 0.653 & 0.346  & 0.614 & 0.255 \\ \midrule
COMET-QE-DA & 0.558 & 0.119  & 0.489 & 0.03  & 0.569 & 0.141  & 0.476 & 0.088 \\
COMET-QE-MQM & 0.545 & 0.132  & 0.552 & 0.106 & 0.574 & 0.195  & 0.574 & 0.148 \\
UniTE-QE & 0.566 & 0.183  & 0.552 & 0.114 & 0.628 & 0.258  & 0.603 & 0.215 \\ \bottomrule
\end{tabular}
\caption{MT metrics for extrinsic Dialogue State Tracking (Multi$^2$WoZ) using an English-trained state tracker. Good/Bad are the number of examples in the respective labels (Not breakdown/Breakdown) for the classification task. Reported Macro F1 scores and MCC scores quantify if the metric detects a breakdown for the extrinsic task. Metrics have negligible correlation with the outcomes of the end task.}
\label{tab:dst-main}
\end{table*}

\begin{table*}[t]
\small
\centering
\resizebox{\linewidth}{!}{%
\begin{tabular}{@{}ccccccccccc@{}}
\toprule
Language & ar & de & el & es & hi & ru & th & tr & vi & zh \\
Good / Bad & 592 / 264 & 696 / 169 & 701/ 170 & 721 / 152 & 631 / 241 & 701 / 173 & 539 / 323 & 443 / 389 & 616 / 251 & 606 / 266 \\ \midrule
Random & 0.023 & -0.002 & -0.002 & 0.017 & 0.001 & -0.002 & -0.002 & 0.028 & -0.051 & -0.045 \\ \midrule
BLEU & 0.135 & 0.048 & 0.142 & 0.098 & 0.162 & 0.125 & 0.128 & 0.097 & 0.108 & 0.171 \\
chrF & 0.160 & 0.083 & 0.172 & 0.092 & 0.202 & 0.106 & 0.162 & 0.000 & 0.173 & 0.119 \\ \midrule
BERTScore & 0.139 & 0.076 & 0.173 & 0.051 & 0.209 & 0.131 & 0.121 & 0.046 & 0.173 & 0.148 \\
COMET-DA & 0.193 & 0.122 & 0.194 & 0.086 & 0.187 & 0.111 & 0.125 & 0.108 & 0.124 & 0.120 \\
COMET-MQM & 0.096 & 0.011 & 0.025 & 0.017 & 0.062 & -0.023 & -0.001 & -0.050 & 0.079 & 0.054 \\
UniTE & 0.068 & -0.031 & -0.002 & -0.014 & 0.043 & 0.047 & -0.006 & 0.056 & -0.017 & -0.023 \\ \midrule
COMET-QE-DA & 0.178 & 0.084 & 0.142 & 0.068 & 0.125 & 0.115 & 0.066 & 0.049 & 0.063 & 0.110 \\
COMET-QE-MQM & 0.099 & 0.050 & -0.013 & 0.025 & 0.090 & -0.025 & 0.041 & -0.077 & 0.068 & 0.070 \\
UniTE-QE & 0.065 & -0.031 & 0.012 & -0.008 & 0.035 & 0.069 & 0.073 & 0.056 & -0.009 & -0.069 \\ \bottomrule
\end{tabular}%
}
\caption{MCC values for different metrics for extrinsic task of Extractive Question Answering (XQuaD dataset) where the model is trained on English. Good/Bad are the number of examples in the respective labels (Not breakdown/Breakdown) for the classification task. Metrics have poor performance on the classification task as a majority report MCC < 0.3}
\label{tab:xquad-main}
\end{table*}
\begin{table*}
\small
\centering
\resizebox{\linewidth}{!}{%

\begin{tabular}{@{}cllllllllll@{}}
\toprule
Method & ar & de & el & es & hi & ru & th & tr & vi & zh \\ \midrule
Good / Bad & 592 / 264 & 696 / 169 & 701/ 170 & 721 / 152 & 631 / 241 & 701 / 173 & 539 / 323 & 443 / 389 & 616 / 251 & 606 / 266 \\ \midrule
Random & 0.508 & 0.525 & 0.512 & 0.492 & 0.489 & 0.505 & 0.490 & 0.468 & 0.473 & 0.498 \\ \midrule
BLEU & 0.549 & 0.515 & 0.564 & 0.543 & 0.571 & 0.562 & 0.556 & 0.487 & 0.549 & 0.585 \\
chrF & 0.579 & 0.541 & 0.575 & 0.546 & 0.595 & 0.545 & 0.567 & 0.480 & 0.557 & 0.554 \\ \midrule
BERTScore & 0.569 & 0.538 & 0.586 & 0.523 & 0.604 & 0.528 & 0.561 & 0.523 & 0.580 & 0.535 \\
COMET-DA & 0.596 & 0.560 & 0.571 & 0.543 & 0.593 & 0.543 & 0.561 & 0.549 & 0.562 & 0.540 \\
COMET-MQM & 0.535 & 0.351 & 0.307 & 0.225 & 0.361 & 0.365 & 0.330 & 0.429 & 0.509 & 0.453 \\
UniTE & 0.370 & 0.479 & 0.343 & 0.314 & 0.308 & 0.519 & 0.366 & 0.438 & 0.282 & 0.326 \\ \midrule
COMET-QE-DA & 0.575 & 0.534 & 0.559 & 0.530 & 0.550 & 0.544 & 0.532 & 0.474 & 0.530 & 0.495 \\
COMET-QE-MQM & 0.549 & 0.510 & 0.416 & 0.473 & 0.420 & 0.384 & 0.356 & 0.459 & 0.509 & 0.492 \\
UniTE-QE & 0.356 & 0.217 & 0.344 & 0.363 & 0.322 & 0.534 & 0.525 & 0.416 & 0.281 & 0.523 \\
\bottomrule
\end{tabular}}
\caption{macro F1 scores for different metrics for extrinsic task of Extractive Question Answering (XQuAD dataset) where the model is trained on English. Good/Bad are the number of examples in the respective labels (Not breakdown/Breakdown) for the classification task. }
\label{tab:xquad-F1}
\end{table*}
\begin{table*}
\scriptsize
\centering
\resizebox{\textwidth}{!}{%
\begin{tabular}{@{}llllllllllll@{}} 
\toprule
src & tgt & Random & BLEU & chrF & BERTScore & COMET-DA & COMET-MQM & UniTE & COMET-QE-DA & COMET-QE-MQM & UniTE-QE \\ \midrule
\multirow{5}{*}
  {en} & de & 0.465 & 0.492 & 0.500 & 0.45 & 0.436 & 0.465 & 0.469 & 0.511 & 0.474 &0.481 \\

 & fr & 0.440 & 0.487 & 0.519 & 0.467 & 0.473 & 0.491 & 0.525 & 0.489 & 0.525 &0.509 \\

 & pt & 0.466 & 0.676 & 0.659 & 0.614 & 0.555 & 0.609 & 0.4525 & 0.527 & 0.500 & 0.588 \\

 & es & 0.463 & 0.599 & 0.566 & 0.564 & 0.630 & 0.614 & 0.626 & 0.546 & 0.535 &0.574 \\

 & zh & 0.429 & 0.574 & 0.570 & 0.582 & 0.590 & 0.577 & 0.586 & 0.516 & 0.513 & 0.490 \\

\midrule
\multirow{5}{*}
  {de} & en & 0.490 & 0.611 & 0.598 & 0.623 & 0.624 & 0.637 & 0.629 & 0.556 & 0.620 & 0.673 \\

 & fr & 0.409 & 0.523 & 0.539 & 0.515 & 0.595 & 0.613 & 0.608 & 0.592 & 0.522 &0.536 \\

 & pt & 0.462 & 0.592 & 0.641 & 0.638 & 0.684 & 0.683 & 0.619 & 0.645 & 0.619 & 0.580 \\

 & es & 0.479 & 0.605 & 0.621 & 0.569 & 0.666 & 0.631 & 0.684 & 0.596 & 0.576 &0.621 \\

 & zh & 0.468 & 0.614 & 0.670 & 0.571 & 0.614 & 0.553 & 0.581 & 0.524 & 0.532 &0.554 \\

\multirow{5}{*}
  {fr} & en & 0.489 & 0.595 & 0.590 & 0.607 & 0.630 & 0.606 & 0.628 & 0.597 & 0.574 &0.588 \\

 & de & 0.385 & 0.518 & 0.616 & 0.587 & 0.541 & 0.570 & 0.546 & 0.503 & 0.476 &0.542 \\

 & pt & 0.472 & 0.620 & 0.620 & 0.565 & 0.543 & 0.583 & 0.538 & 0.549 & 0.534 & 0.520 \\

 & es & 0.492 & 0.462 & 0.613 & 0.512 & 0.627 & 0.648 & 0.574 & 0.594 & 0.568 &0.573 \\

 & zh & 0.384 & 0.641 & 0.702 & 0.666 & 0.667 & 0.658 & 0.661 & 0.521 & 0.502 &0.575 \\
\midrule
\multirow{5}{*}
{pt} & en & 0.476 & 0.629 & 0.676 & 0.681 & 0.685 & 0.655 & 0.705 & 0.695 & 0.654 &0.526 \\

 & de & 0.438 & 0.550 & 0.575 & 0.577 & 0.586 & 0.594 & 0.481 & 0.608 & 0.569 &0.501 \\

 & fr & 0.458 & 0.546 & 0.603 & 0.488 & 0.599 & 0.495 & 0.574 & 0.574 & 0.545 &0.645 \\

 & es & 0.491 & 0.640 & 0.646 & 0.634 & 0.639 & 0.639 & 0.459 & 0.562 & 0.586 &0.509 \\

 & zh & 0.403 & 0.610 & 0.690 & 0.551 & 0.580 & 0.511 & 0.621 & 0.621 & 0.492 &0.591 \\

 \midrule
\multirow{5}{*}
 {es} & en & 0.455 & 0.530 & 0.561 & 0.566 & 0.605 & 0.601 & 0.600 & 0.544 & 0.564 &0.529 \\

 & de & 0.455 & 0.530 & 0.546 & 0.587 & 0.540 & 0.521 & 0.584 & 0.49 & 0.486 &0.513 \\

 & fr & 0.453 & 0.542 & 0.531 & 0.606 & 0.564 & 0.568 & 0.584 & 0.569 & 0.560 &0.556 \\

 & pt & 0.500 & 0.506 & 0.561 & 0.579 & 0.554 & 0.564 & 0.529 & 0.561 & 0.566 &0.581 \\

 & zh & 0.374 & 0.562 & 0.644 & 0.562 & 0.627 & 0.587 & 0.687 & 0.524 & 0.478 &0.662 \\
 \midrule
\multirow{5}{*}
{es} & en & 0.455 & 0.530 & 0.561 & 0.566 & 0.605 & 0.601 & 0.600 & 0.544 & 0.564 &0.529 \\

 & de & 0.455 & 0.530 & 0.546 & 0.587 & 0.540 & 0.521 & 0.584 & 0.490 & 0.486 &0.513 \\

 & fr & 0.453 & 0.542 & 0.531 & 0.606 & 0.564 & 0.568 & 0.584 & 0.569 & 0.560 &0.556 \\

 & pt & 0.500 & 0.506 & 0.561 & 0.579 & 0.554 & 0.564 & 0.529 & 0.561 & 0.566 &0.581 \\

 & zh & 0.374 & 0.562 & 0.644 & 0.562 & 0.627 & 0.587 & 0.687 & 0.524 & 0.478 &0.662 \\
 \bottomrule
\end{tabular}
}
\caption{MT Metric performance on F1 for extrinsic semantic parsing (MultiATIS++SQL) with the parser trained in src language.}
\label{tab:parsing-f1}
\end{table*}
\begin{table*}
\scriptsize
\centering
\resizebox{\textwidth}{!}{%
\begin{tabular}{@{}llllllllllll@{}} \toprule
src & tgt & Random & BLEU & chrF & BERTScore & COMET-DA & COMET-MQM & UniTE & COMET-QE-DA & COMET-QE-MQM & UniTE-QE \\ \midrule
\multirow{5}{*}{en} & de & 0.012 & 0.008 & 0.016 & -0.096 & -0.122 & -0.000 &-0.06& 0.025 & -0.021 &-0.027 \\

 & fr & -0.043 & -0.024 & 0.039 & -0.066 & -0.020 & -0.001 &0.050& -0.021 & -0.021 &0.017 \\

 & pt & -0.067 & 0.353 & 0.328 & 0.231 & 0.201 & 0.228 &0.114& 0.089 & 0.209 &0.187 \\

 & es & 0.002 & 0.203 & 0.133 & 0.152 & 0.279 & 0.229 &0.252& 0.110 & 0.107 &0.166 \\

 & zh & -0.090 & 0.152 & 0.146 & 0.173 & 0.187 & 0.188 &0.172& 0.060 & 0.035 &0.078 \\ \midrule

\multirow{5}{*}{de} & en & -0.003 & 0.226 & 0.210 & 0.251 & 0.263 & 0.328 &0.303& 0.161 & 0.250 &0.349 \\

 & fr & -0.007 & 0.046 & 0.078 & 0.033 & 0.196 & 0.226 &0.243& 0.185 & 0.044 &0.078 \\

 & pt & -0.070 & 0.184 & 0.300 & 0.312 & 0.394 & 0.406 &0.302& 0.331 & 0.295 &0.206 \\

 & es & -0.035 & 0.230 & 0.242 & 0.200 & 0.332 & 0.264 &0.370 & 0.206 & 0.181 &0.256 \\

 & zh & -0.063 & 0.241 & 0.340 & 0.150 & 0.242 & 0.124 &0.258& 0.054 & 0.088 &0.112 \\ \midrule

\multirow{5}{*}{fr} & en & 0.006 & 0.194 & 0.182 & 0.220 & 0.269 & 0.229 &0.262& 0.195 & 0.148 &0.178 \\

 & de & -0.087 & 0.099 & 0.237 & 0.180 & 0.105 & 0.155 &0.125& 0.026 & -0.043 &0.086 \\

 & pt & -0.023 & 0.242 & 0.240 & 0.177 & 0.133 & 0.170 &0.117& 0.100 & 0.115 &0.106 \\

 & es & -0.015 & 0.053 & 0.233 & 0.118 & 0.283 & 0.300 &0.151& 0.229 & 0.177 &0.153 \\

 & zh & -0.116 & 0.311 & 0.413 & 0.373 & 0.365 & 0.347 &0.390 & 0.143 & 0.051 &0.248 \\ \midrule

\multirow{5}{*}{pt} & en & 0.013 & 0.315 & 0.365 & 0.378 & 0.372 & 0.320 &0.414& 0.402 & 0.310 &0.175 \\

 & de & -0.093 & 0.112 & 0.181 & 0.159 & 0.188 & 0.190 &0.216& 0.183 & 0.150 &0.007 \\

 & fr & 0.013 & 0.100 & 0.222 & 0.061 & 0.218 & 0.030 &0.155& 0.053 & 0.090 &0.291 \\

 & es & 0.009 & 0.286 & 0.293 & 0.278 & 0.278 & 0.288 &0.142& 0.076 & 0.243 &0.025 \\

 & zh & 0.061 & 0.221 & 0.449 & 0.253 & 0.161 & 0.048 &0.242& 0.000 & -0.011 &0.212 \\ \midrule

\multirow{5}{*}{es} & en & -0.063 & 0.080 & 0.179 & 0.136 & 0.214 & 0.208 & 0.200 & 0.095 & 0.128 &0.058 \\

 & de & -0.075 & 0.092 & 0.169 & 0.175 & 0.082 & 0.047 &0.186& -0.013 & -0.024 &0.033 \\

 & fr & -0.065 & 0.140 & 0.118 & 0.214 & 0.129 & 0.140 &0.196& 0.150 & 0.124 &0.112 \\

 & pt & 0.014 & 0.012 & 0.144 & 0.169 & 0.148 & 0.143 &0.110& 0.160 & 0.133 &0.166 \\

 & zh & -0.005 & 0.148 & 0.289 & 0.154 & 0.254 & 0.173 &0.393& 0.102 & 0.000 &0.363 \\ \midrule

\multirow{5}{*}{zh} & en & -0.034 & 0.283 & 0.218 & 0.252 & 0.302 & 0.290 &0.333& 0.264 & 0.324 &0.232 \\

 & de & 0.008 & 0.260 & 0.274 & 0.302 & 0.314 & 0.347 &0.273& 0.139 & 0.199 &0.169 \\

 & fr & -0.045 & 0.204 & 0.238 & 0.343 & 0.330 & 0.247 &0.328& 0.222 & 0.259 &0.287 \\

 & pt & -0.130 & 0.264 & 0.357 & 0.430 & 0.327 & 0.295 &0.307& 0.171 & 0.205 &0.134 \\

 & es & -0.015 & 0.340 & 0.375 & 0.446 & 0.407 & 0.417 &0.213& 0.139 & 0.229 &0.211 \\ 
 \bottomrule
\end{tabular}
}
\caption{MT Metric performance on MCC for the classification task with extrinsic semantic parsing (MultiATIS++SQL) with the parser trained in src language. }
\label{tab:parsing-mcc}
\end{table*}
\end{document}